# Circadian Modulation of Semantic Exploration in Social Media Language

Vuong Hung Truong[1], Mariana Gabrielle Cangco Reyes[2], Masatoshi Koizumi[3], Jihwan Myung[1,4,*]

[1] Graduate Institute of Mind, Brain and Consciousness (GIMBC), Taipei Medical University, New Taipei City 235, Taiwan
[2] International Degree Program in Climate Change and Sustainable Development, National Taiwan University, Taipei 106319, Taiwan
[3] Department of Linguistics, Graduate School of Arts and Letters, Tohoku University, Sendai, Miyagi 980-8577, Japan
[4] Graduate Institute of Medical Sciences, College of Medicine, Taipei Medical University, Taipei 110, Taiwan

* Correspondence: Jihwan Myung, jihwan@tmu.edu.tw

## Abstract

Human cognition exhibits strong circadian modulation, yet its influence on high-dimensional semantic behavior remains poorly understood. Using large-scale Reddit data, we quantify time-of-day variation in language use by embedding text into a pretrained transformer model and measuring semantic entropy as an index of linguistic exploration-exploitation, for which we show a robust circadian rhythmicity that could be entrained by seasonal light cues. Distinguishing between local and global semantic entropy reveals a systematic temporal dissociation: local semantic exploration peaks in the morning, reflecting broader exploration of semantic space, whereas global semantic diversity peaks later in the day as submissions accumulate around already established topics, consistent with "rich-get-richer" dynamics. These patterns are not explained by sentiment or affective valence, indicating that semantic exploration captures a cognitive dimension distinct from mood. The observed temporal structure aligns with known diurnal patterns in neuromodulatory systems, suggesting that biological circadian rhythms extend to the semantic domain.

**Keywords:** Circadian rhythms, semantic embedding, semantic entropy, social media, psycholinguistics

**Running title:** Circadian Semantic Exploration

## Highlights

- Circadian semantic exploration is a robust and cross-national phenomenon.
- Seasonal light cues may entrain semantic exploration.
- Semantic exploration is dissociable from sentiment and affect.
- Semantic networks follow "rich-get-richer" clustering over the day.

## Graphical abstract

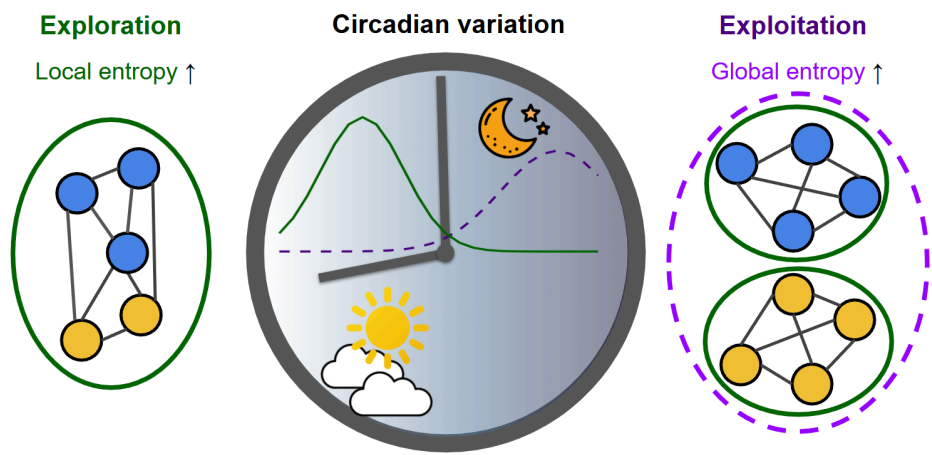

Social media language use reveals circadian variation in semantic exploration and exploitation, quantified by local and global entropy. Nodes represent semantic embeddings, and edges connect k-nearest neighbors. The green outline indicates the semantic volume proportional to local entropy, while the dashed violet outline indicates increased global entropy during exploitation.

# 1. Introduction

Daily rhythms in life, such as sleep and wakefulness, are governed by circadian clocks. These clocks are fundamentally biological, based on molecular gene expression, and regulate the rhythmic homeostasis of mammalian physiology. Human cognition exhibits pronounced circadian modulation, influencing attention, affect, and decision-making. Large-scale analyses of social media data have revealed daily rhythms in emotional expression and word usage, suggesting systematic time-of-day variation in language production. These lexical patterns reflect cross-cultural diurnal mood rhythms, with positive affect (PA) words dominating in the morning and rebounding in the evening, whereas negative affect (NA) words increase progressively toward night (Golder & Macy, 2011). While diurnal fluctuations in language use can reflect various social, cultural, and environmental factors, they are also driven by endogenous circadian rhythms. Subsequent analyses of Twitter content further partitioned the NA vocabulary to show that even within the same affect category, certain NA words exhibit stable circadian patterns, while others are more sensitive to environmental factors such as seasonal light onset (Dzogang et al., 2017). The COVID-19 lockdowns provided a unique opportunity to disentangle social influences, revealing that mood-related language use preserved its circadian rhythmicity even during isolation (Wang et al., 2024).

However, whether these fluctuations reflect changes in semantic exploration, a core component of cognitive flexibility, remains unclear. Semantic exploration can be understood as the degree to which individuals "travel" through ideas and words when they communicate, or whether they stick to familiar expressions or branch out into new, creative ways of saying things. In cognitive science, this is closely related to the exploration-exploitation trade-off principle, which describes how agents use known knowledge (exploitation) versus trying new possibilities (exploration) to maximize outcomes (Cohen et al., 2007; Daw et al., 2006). To quantify semantic exploration across time, we derived two complementary measures from high-dimensional text embeddings: local semantic entropy and global semantic entropy. Both are inspired by Shannon's information entropy (Shannon, 1948). Local entropy captures the variability of individual submissions within a given country and hour, reflecting how much each post diverges from other posts in the same group, analogous to the "exploration" side of the exploration-exploitation trade-off. Global entropy, by contrast, measures the overall dispersion of all posts in a country at a given hour, capturing the breadth of semantic space covered by the population as a whole. We compute it using the determinant of the covariance matrix of embeddings, which is a continuous, multivariate analogue of Shannon entropy, known as differential entropy (Cover & Thomas, 2005).

Using these measures, we analyze a large recent Reddit corpus and compare the resulting circadian patterns to previously reported word usage rhythms by extracting published data from Dzogang et al. (2017). This comparison allows us to relate embedding-based semantic dynamics to established circadian structure in lexical usage. By separating local from global semantic structure, we dissociate individual-level exploratory dynamics from population-level heterogeneity and extend circadian analyses of language beyond word frequency and affect.

## 2. Methods

**2.1. Semantic exploration and sentiment in Reddit submissions**

We analyzed publicly available Reddit submissions retrieved via the Pushshift archive API (all submissions from January 2024 to December 2024; Baumgartner et al., 2020). Only publicly accessible content was used, and all analyses were conducted on anonymized text. For each submission, textual content was constructed by concatenating the post title and selftext, with missing values replaced by empty strings. Posts with empty, deleted, or removed content were excluded from further analysis. Each submission was assigned a circadian time based on its inferred local posting hour, discretized into 1-h bins (0-23). To focus on language that likely reflects genuine human expression, we applied several filtering steps. Submissions marked as NSFW were removed, as were posts associated with advertisements or promotions. Automated and bot-generated content was excluded by removing posts from accounts with bot-related identifiers and platform-level ad flags. Finally, analyses were restricted to posts written in English, detected using automatic language identification.

**2.2. Geographic attribution**

We inferred submission-level geographic context, assigning each post to a country (and when possible, a city) using a hierarchical strategy: (1) URL-based inference (Top-level domains, e.g., `.uk, .de, .jp`; well-known national news outlets, e.g., `nytimes.com`, `bbc.co.uk`); (2) Subreddit-based inference. If no location was assigned from the URL, we next checked whether the subreddit name directly corresponded to a known city (e.g., r/london, r/berlin); (3) Author flair-based inference; and finally, (4) Text-based inference using named-entity recognition (NER). For each city-country pair, geographic coordinates (latitude and longitude) were obtained via forward geocoding using the `Nominatim` (OpenStreetMap) service accessed through the `geopy` library. Submissions for which no reliable geographical information was detected at any level were excluded from further analysis.

Cleaning and preprocessing were conducted in R 4.4.2. City-level data were harmonized and aggregated using publicly available datasets for the United Kingdom (UK), United States (US), Canada, and India (Kaggle and SimpleMaps). Reddit posts were spatially aggregated to city and administrative-region levels. Spatial visualization and graphing were performed using ArcGIS Pro 3.6.0.

**2.3. Local time conversion, daylight saving correction, and sunrise-sunset estimation**

All timestamps in the raw Reddit data are recorded in Coordinated Universal Time (UTC). To analyze circadian patterns, we converted these timestamps to local clock time for each submission based on its inferred geographic location. This conversion also accounted for daylight saving time (DST). Submissions for which a reliable time zone could not be determined were excluded.

Sunrise and sunset times were computed using an astronomical solar position model implemented in the `Astral` library. To ensure consistency with DST transitions, all sunrise and sunset times were computed in local civil time using the city-specific time zone information. Because our analyses were conducted at a monthly resolution, we stabilized seasonal variation by computing sunrise and sunset on the 15th day of each month, which serves as a representative midpoint of the month. The resulting sunrise and sunset times were then aligned with local-hour bins and overlaid on circadian heatmaps as reference markers, allowing direct comparison between linguistic entropy rhythms and natural light–dark cycles across countries and seasons.

### 2.4. Sentiment estimation

Sentiment was quantified using the VADER sentiment analyzer, a lexicon-and-rule-based model specifically designed for quantifying social media text's sentiment (https://github.com/cjhutto/vaderSentiment) (Hutto & Gilbert, 2014). For each submission, we computed the compound sentiment score by integrating positive, negative, and neutral valence into a single normalized metric, which ranges from -1 (strongly negative) to +1 (strongly positive).

To validate our sentiment estimates against established circadian patterns of emotional expression, we used published lexical emotion variability profiles from Dzogang et al. (2017). Specifically, hourly standard deviation (SD) time series for positive emotion (posemo) and sadness word usage were digitized directly from Figures 8b and 9b of the original publication using WebPlotDigitizer 5.2 (https://automeris.io/wpd). In the original study, these indicators were derived from Twitter data using predefined LIWC word lists (Tausczik & Pennebaker, 2010), with emotion signals computed as the average standardized time series across words in each category after restricting analyses to the 50% most frequent terms to mitigate Zipfian frequency effects and reduce noise from low-frequency words. We did not reprocess or reanalyze the underlying Twitter corpus; instead, the digitized SD profiles were used as external reference signals reflecting population-level variability in emotional word usage across the day in the UK.

### 2.5. Local and global semantic entropy

Text was embedded using a pretrained transformer model (`sentence-transformers/all-mpnet-base-v2`), producing normalized sentence embeddings for each post.

Semantic exploration was quantified for each post using a k-nearest-neighbor (kNN) entropy proxy in embedding space. For a given circadian hour, let $X = \{x_1, x_2, ..., x_N\} \subset \mathbb{R}^d$ be the set of normalized text embeddings. For each text embedding $x_i$, entropy was estimated as:

$$H_{local}(x_i) = \log(r_k(x_i))$$

where $r_k(x_i)$ denotes the distance from the post $x_i$ to its $k$-th nearest neighbor (we used $k = 10$) in embedding space. This approach captures how semantically distinct a post is relative to other posts in the same circadian bin.

To quantify population-level semantic diversity, we computed the differential entropy of the distribution of sentence embeddings within each circadian hour. For each hour, embeddings were treated as samples from a multivariate Gaussian distribution, and entropy was estimated using the closed-form expression:

$$H_{global} = \frac{1}{2}\log((2\pi e)^d \det(\Sigma))$$

where $d$ is the embedding dimensionality, and $\Sigma$ is the covariance matrix of the embeddings. This measure captures the overall semantic spread or volume occupied by posts in embedding space, with higher values indicating greater diversity of discussed topics.

### 2.6. Statistical analyses

To reduce the influence of extreme values, we excluded outliers from the local semantic entropy prior to aggregation and cosinor fitting. Outliers were defined using the interquartile range (IQR) method: values below $Q1 - 1.5 \cdot IQR$ or above $Q3 + 1.5 \cdot IQR$ were removed, where $Q1$ and $Q3$ are the 25th and 75th percentiles, respectively. After exclusion, the remaining data were used to compute hourly means, standard errors, heatmaps, and cosinor fits. To quantify circadian rhythmicity, we applied a single-component 24-h cosinor model. For each country, individual datapoints were fit to the model:

$$y(t) = M + \beta_{cos}\cos(\omega t) + \beta_{sin}\sin(\omega t) + \epsilon(t)$$

where $M$ indicates the mesor and $\epsilon(t)$ is the residual error. Amplitude $A$ and acrophase $\phi$ were derived from the fitted coefficients as

$$A = \sqrt{\beta_{cos}^2 + \beta_{sin}^2}, \quad \phi = \frac{1}{\omega}atan2(\beta_{sin}, \beta_{cos}).$$

Model significance was assessed using a likelihood-ratio test, comparing the full cosinor model against a null model that contained only the mean. The coefficient of determination ($r^2$) was also reported as a measure of fit quality. All analyses were performed using Python's `statsmodels`.

Pearson correlation coefficients (*r*) were used to assess associations between continuous variables, and two-sample *t*-tests were applied for group comparisons. Multiple comparisons were controlled using the false discovery rate (FDR) procedure (Benjamini & Hochberg, 1995).

## 3. Results

### 3.1. Circadian local semantic entropy

We analyzed 2,921,362 Reddit submissions from the top 4 countries with the most submissions (United States: 1,739,146 posts; India: 458,776 posts; United Kingdom: 385,138 posts; Canada: 338,302 posts), collected from January 2024 to December 2024 (**Figure 1**). All four countries exhibited significant 24-h rhythmicity in local semantic entropy ($P_{fdr} < 0.001$). The amplitude of the local semantic entropy rhythm, representing the extent of variation in semantic exploration across the day, ranged from 0.036 in the UK to 0.074 in India. The timing of peak semantic exploration (acrophase) occurred in the early morning hours: the US at 3.1 h, India at 5.0 h, the UK at 4.2 h, and Canada at 3.2 h. The cosinor model accounted for a substantial proportion of the hourly variation in semantic entropy, with $r^2$ values ranging from 0.73 to 0.78 across countries (**Table 1**). Conceptual illustrations of local and global semantic entropy dynamics are provided in **Figure S1**, and the global distribution of submissions across all countries can be visualized in the world map shown in **Figure S2**.

**Table 1. Cosinor analysis of local semantic entropy**

| Country | Amplitude | Acrophase (h) | $r^2$ | $P_{fdr}$ |
|---|---|---|---|---|
| United States | 0.067 | 3.1 | 0.75 | 7.79E-08 |
| India | 0.074 | 5.0 | 0.74 | 1.04E-07 |
| United Kingdom | 0.036 | 4.2 | 0.78 | 3.77E-08 |
| Canada | 0.063 | 3.2 | 0.78 | 3.77E-08 |

*Notes:* Results of single-component 24-h cosinor fits to hourly local semantic entropy for each country. Amplitude quantifies the strength of circadian modulation, and acrophase indicates the timing of peak semantic exploration (in local time). $r^2$ denotes the proportion of variance explained by the cosinor model. $P_{fdr}$ values were assessed using likelihood-ratio tests against a mean-only null model and corrected for multiple comparisons using FDR. Note that the cosinor model was fitted to a single mean entropy value at each time point, rather than individual submissions' entropy, to avoid the "*N*-Problem" (large sample size issue).

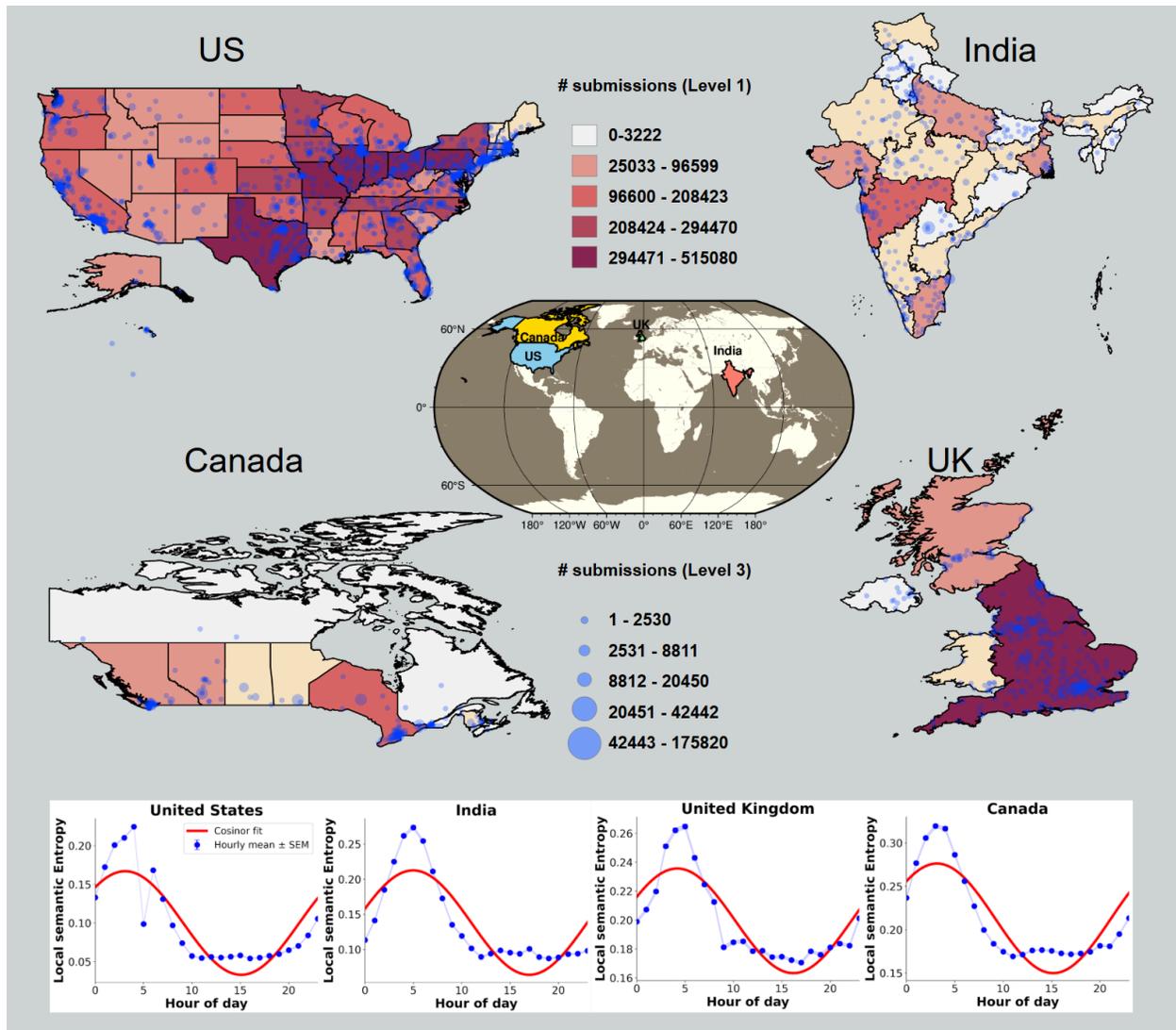

**Figure 1. Geographic distribution of Reddit submissions and circadian local semantic entropy across the top 4 countries with the most submissions. Top panels** show the spatial distribution of Reddit submissions aggregated at the Nomenclature of Territorial Units for Statistics (NUTS) level 1 and level 3 for the United States, India, the United Kingdom, and Canada, with color intensity indicating the total number of submissions at NUTS level 1, and blue circles of different sizes representing the total number of submissions at NUTS level 3. A world map inset displays the geographic locations of the four countries using a Robinson projection with preserved relative country scales. **Bottom panels** show circadian profiles of local semantic entropy for each country. Blue points represent hourly mean local semantic entropy (± SEM), and red curves denote the fitted 24-h cosinor models. Despite substantial differences in geography, population distribution, and posting volume, all countries exhibit a robust circadian rhythm in local semantic entropy, with peaks occurring in the early morning hours.

## 3.2. Seasonal variation in local semantic entropy

Next, we examined whether circadian semantic exploration was influenced by photoperiod. We computed the average local semantic entropy for each month and visualized seasonal changes using heatmaps (**Figure 2A**). Correlation analyses indicated that photoperiod was associated with shifts in the timing of semantic exploration. In the United States, the circadian trough of semantic exploration was positively correlated with the timing of sunset ($r$ = 0.68, $P_{fdr}$ = 0.03; **Figure 2C**); however, the relationship between the peak of semantic exploration and the timing of sunrise was not significant ($r$ = -0.04, $P_{fdr}$ = 0.901; **Figure 2B**). In other countries, although correlations did not reach significance when averaged across entire months, seasonal alignment was nonetheless apparent upon visual inspection of the heatmaps for specific periods (e.g., from March to October in Canada). Together, these findings suggest that seasonal changes in day length may entrain the phase of semantic activity.

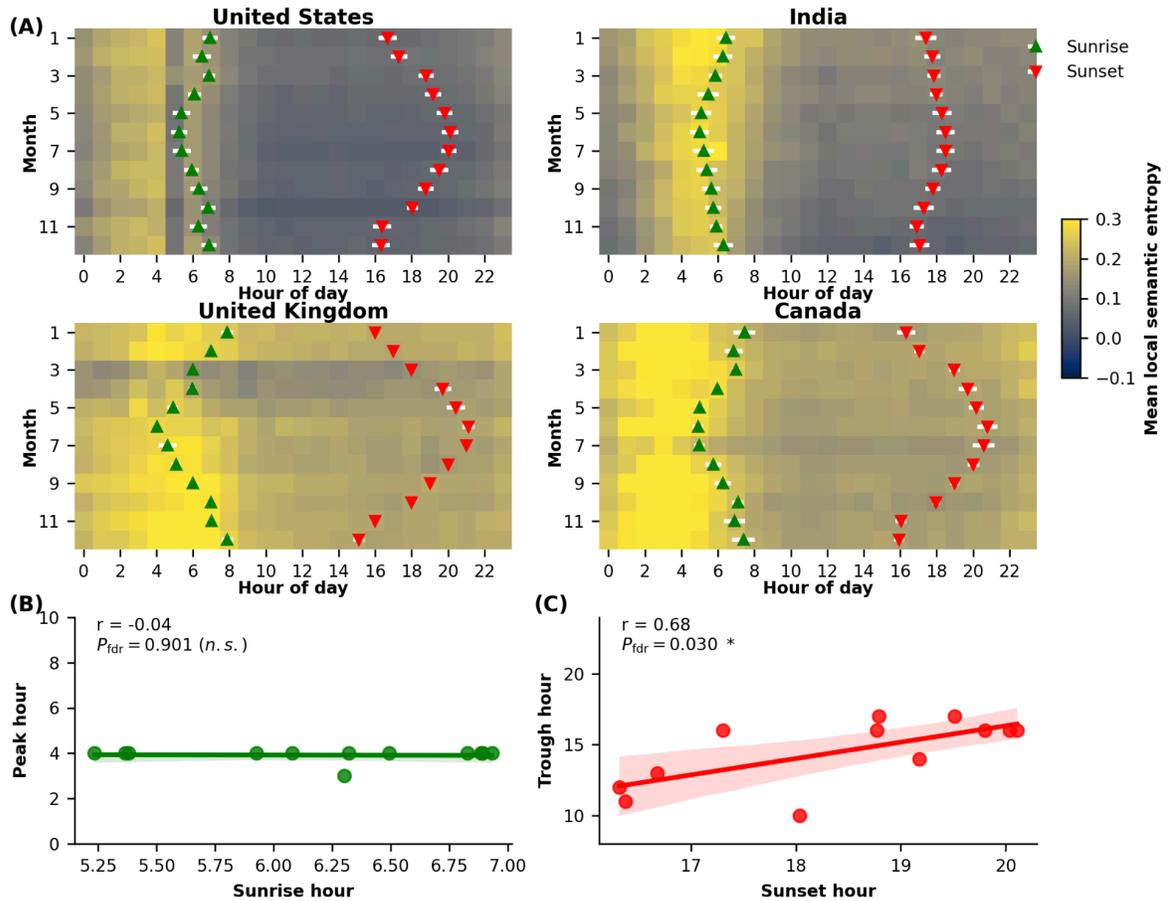

**Figure 2. Seasonal entrainment of local semantic entropy across countries. (A)** Heatmaps show mean local semantic entropy for the US, India, the UK, and Canada. The colorbar indicates mean local semantic entropy (scale at right). Green upward triangles mark monthly mean sunrise times and red downward triangles mark monthly mean sunset times for each country, with standard deviation (SD) indicated in white. **(B)** Correlation between sunrise hour and the morning peak hour of local semantic entropy. **(C)** Correlation between sunset hour and the evening trough hour of local semantic entropy. The solid line indicates the linear fit, and the shaded area represents the 95% confidence interval (CI).

### 3.3. Correlation between semantic exploration and sentiment

If semantic exploration is modulated by affective state, such modulation should be reflected in sentiment scores. We therefore visualized seasonal-circadian heatmaps of average sentiment and observed significant correlations with semantic entropy heatmaps (**Figure 3A**). Notably, the direction of this relationship differed across countries (US: $r$ = -0.543, $P_{fdr}$ < 0.001; Canada: $r$ = -0.259, $P_{fdr}$ < 0.001; UK: $r$ = 0.219, $P_{fdr}$ < 0.001; India: $r$ = 0.134, $P_{fdr}$ = 0.02; **Figure 3B**). Specifically, semantic exploration was negatively associated with sentiment in the United States and Canada, but positively associated with sentiment in the United Kingdom and India.

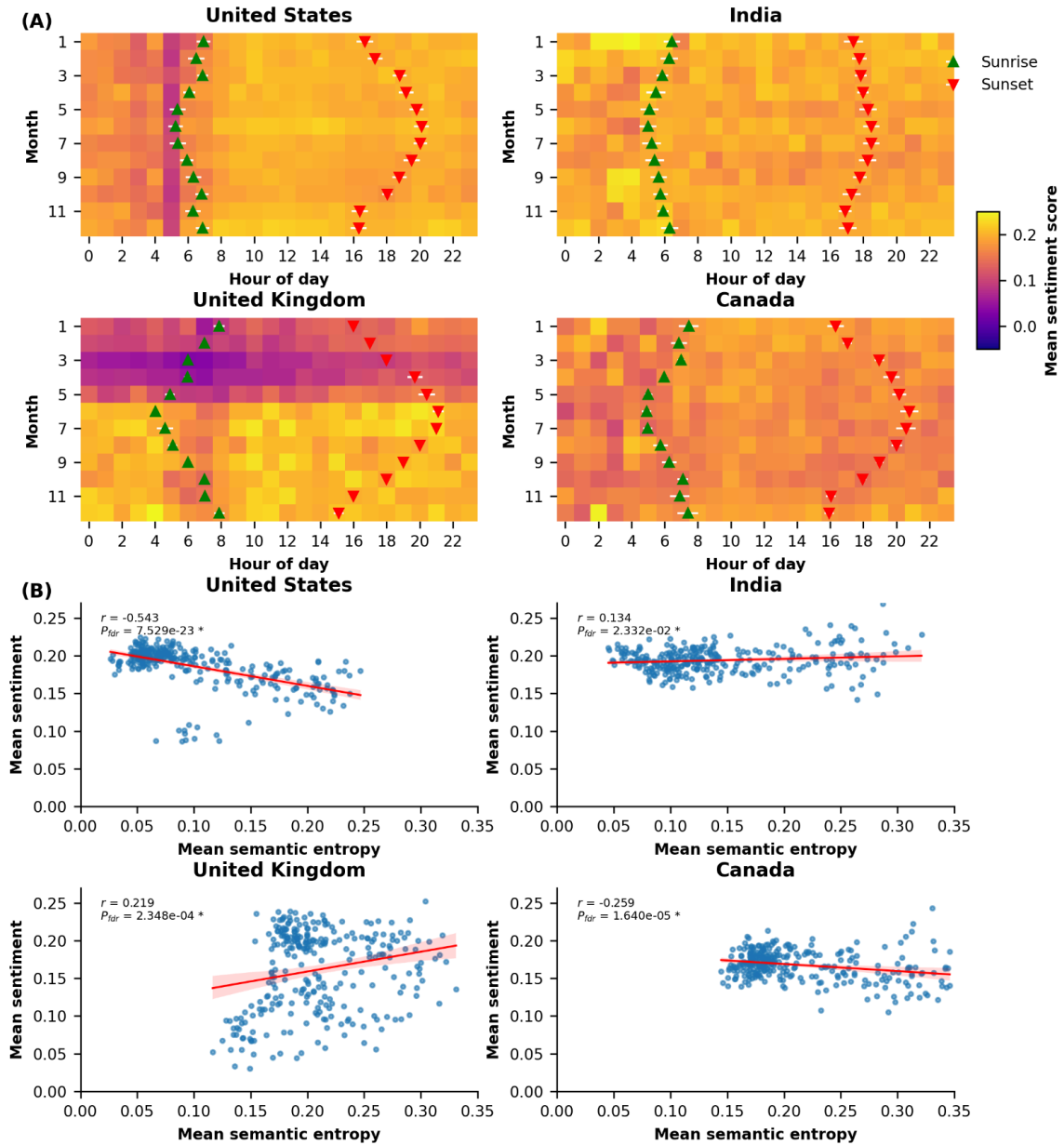

**Figure 3. Seasonal-circadian structure of sentiment score and its relationship to semantic exploration across countries. (A)** Seasonal heatmaps of mean compound sentiment scores across 24 h for the US, India, the UK, and Canada.The colorbar indicates mean sentiment score (scale at right). Green upward triangles mark monthly mean sunrise times, and red downward triangles mark monthly mean sunset times for each country, with SD indicated in white. Note that the sentiment scores did not show a robust circadian rhythm across countries. The UK showed sentiment changes during the wintertime (January to May). **(B)** Correlation between seasonal-circadian local semantic entropy heatmaps and seasonal-circadian sentiment heatmaps for each country. The solid lines indicate linear fits, and the shaded areas represent the 95% CI. Across countries, correlations between

semantic entropy and sentiment are significant but vary in direction and magnitude, indicating that semantic exploration is only weakly and inconsistently related to affective valence.

### 3.4. Mean and variability in circadian fluctuations of word usage

Data from Dzogang et al. (2017; Figures 8b and 9b) supported the validity of our semantic entropy measure, as both showed elevated variability during early-morning hours. Consistent with this observation, the circadian profile of local semantic entropy closely aligned with the standard deviation of sadness-related word usage ($r = 0.74$, $P_{fdr} < 0.001$; **Figure 4A**), but not with the standard deviation of positive emotion (posemo) usage ($r = 0.18$, $P_{fdr} = 0.394$; **Figure 4A**).

In contrast, sentiment scores themselves exhibited weak or absent circadian rhythmicity (**Figure 3A**), suggesting that emotional expression is complex and multifaceted rather than captured by a single rhythmic dimension. Notably, when compared with the Twitter-derived mood indicator reported by Dzogang et al. (2017), our Reddit sentiment score showed a significant positive correlation with average posemo usage ($r = 0.57$, $P_{fdr} = 0.007$; **Figure 4B**), while its correlation with average sadness usage was non-significant ($r = -0.10$, $P_{fdr} = 0.656$; **Figure 4C**). Together, these findings suggest that semantic entropy may be more sensitive to fluctuations in emotional variability, particularly negative affect, than to mean sentiment levels.

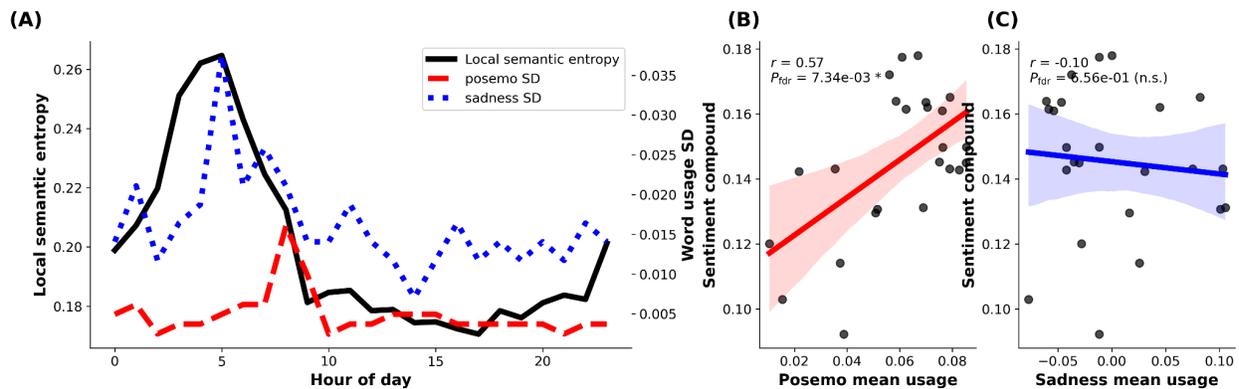

**Figure 4. Correlation between local semantic entropy and affective word-usage variability.** **(A)** Diurnal profile of local semantic entropy from our analysis (black solid line) compared with digitized SD of positive emotion (posemo; red dashed line) and sadness (blue dotted line) word usage from Figures 8b and 9b of Dzogang et al. (2017), extracted using WebPlotDigitizer. Local semantic entropy shows elevated values in the early morning hours, paralleling the high variability in sadness word usage reported previously, and to a lesser extent with posemo SD, although the posemo SD peak was more delayed. **(B)** Mean sentiment score as a function of posemo mean usage, showing a significant positive association. (**C**) Mean sentiment score as a function of sadness mean usage, showing no significant relationship.

### 3.5. Circadian global semantic entropy

In addition to local semantic entropy, we observed a pronounced circadian rhythmicity in global semantic entropy ($P_{fdr}$ <0.001; $r^2$ ranged from 0.56 to 0.78; **Table 2**). Unlike local entropy, which reflects semantic exploration at the level of individual submissions, global entropy captures the overall population-level diversity of submission content. As expected, increases in posting activity over the course of the day were accompanied by an expansion of the semantic embedding space, resulting in strong positive correlations between global semantic entropy and total submission volume (US: $r$ = 0.849, $P_{fdr}$ < 0.001; India: $r$ = 0.829, $P_{fdr}$ < 0.001; UK: $r$ = 0.762, $P_{fdr}$ < 0.001; Canada: $r$ = 0.902, $P_{fdr}$ < 0.001). Interestingly, peaks in local semantic entropy occurred earlier in the day (**Figure 1**), preceding subsequent increases in global semantic diversity later in the day (**Figure 5A**). The full circadian-seasonal heatmaps of global semantic entropy and submission activity are presented in **Figure S3**.

To quantify how additional posts contribute to global semantic diversity, we computed the marginal change in global semantic entropy as a function of cumulative posting volume across countries. For each country, posts were aggregated by month and local hour, and cumulative sums of posting counts and the magnitude of global entropy were calculated. The marginal gain in entropy was defined as the gradient of cumulative entropy with respect to cumulative posts. To examine non-linear effects, we split the cumulative sequence into early (first 15% of cumulative volume) and late segments (the remaining 85%) and fit separate log-log regressions to the marginal gain in each segment. Across all countries, we observed a substantial decline in marginal entropy change, with the mean gain in the late segment reduced from 66.5% (UK) to 76.9% (India) relative to the early segment. This indicates that additional posts contribute progressively less new semantic content, consistent with diminishing returns. The fitted slopes further illustrate the difference in scaling between early and late posting activity, highlighting strong non-linear effects in the accumulation of collective semantic diversity (**Figure 5B**). Additionally, the relationship follows an approximate power-law distribution.

To further characterize the temporal structure underlying the observed entropy dynamics, we examined the organization and growth of semantic clusters over the circadian cycle. Using US data from January as an illustrative case, we visualized the evolution of semantic clusters across the day (**Figure 6A**). Early in the day, submissions were broadly distributed across the semantic embedding space, reflecting widespread semantic exploration. In contrast, later hours were dominated by accumulation within already established clusters, indicating continued engagement with existing topics rather than expansion into novel semantic regions. This qualitative shift mirrors the temporal dissociation between early peaks in local semantic entropy and later increases in global semantic diversity.

Consistent with this observation, the distribution of cluster sizes exhibited a scale-free organization (**Figure 6B**), with a small number of large clusters accounting for the majority of submissions. The fitted log-log slope revealed an approximate power-law relationship between cluster size and frequency ($P(s) \sim s^{-1.60}$). Examination of cumulative cluster growth over the full 24-h cycle further demonstrated that most posting activity accrued within these dominant clusters, whereas smaller clusters accumulated relatively few submissions (**Figure 6C**).

**Table 2. Cosinor analysis of global semantic entropy**

| Country | Amplitude | Acrophase (h) | $r^2$ | $P_{fdr}$ |
|---|---|---|---|---|
| United States | 1.150 | 14.78 | 0.661 | 3.07E-06 |
| India | 1.067 | 16.9 | 0.570 | 4.03E-05 |
| United Kingdom | 1.245 | 17.4 | 0.775 | 6.56E-08 |
| Canada | 1.155 | 15.1 | 0.668 | 3.07E-06 |

*Notes:* Results of single-component 24-h cosinor fits to hourly global semantic entropy for each country. Amplitude quantifies the strength of circadian modulation, and acrophase indicates the timing of peak semantic exploitation (in local time). $r^2$ denotes the proportion of variance explained by the cosinor model. $P_{fdr}$ values were assessed using likelihood-ratio tests against a mean-only null model and corrected for multiple comparisons using FDR.

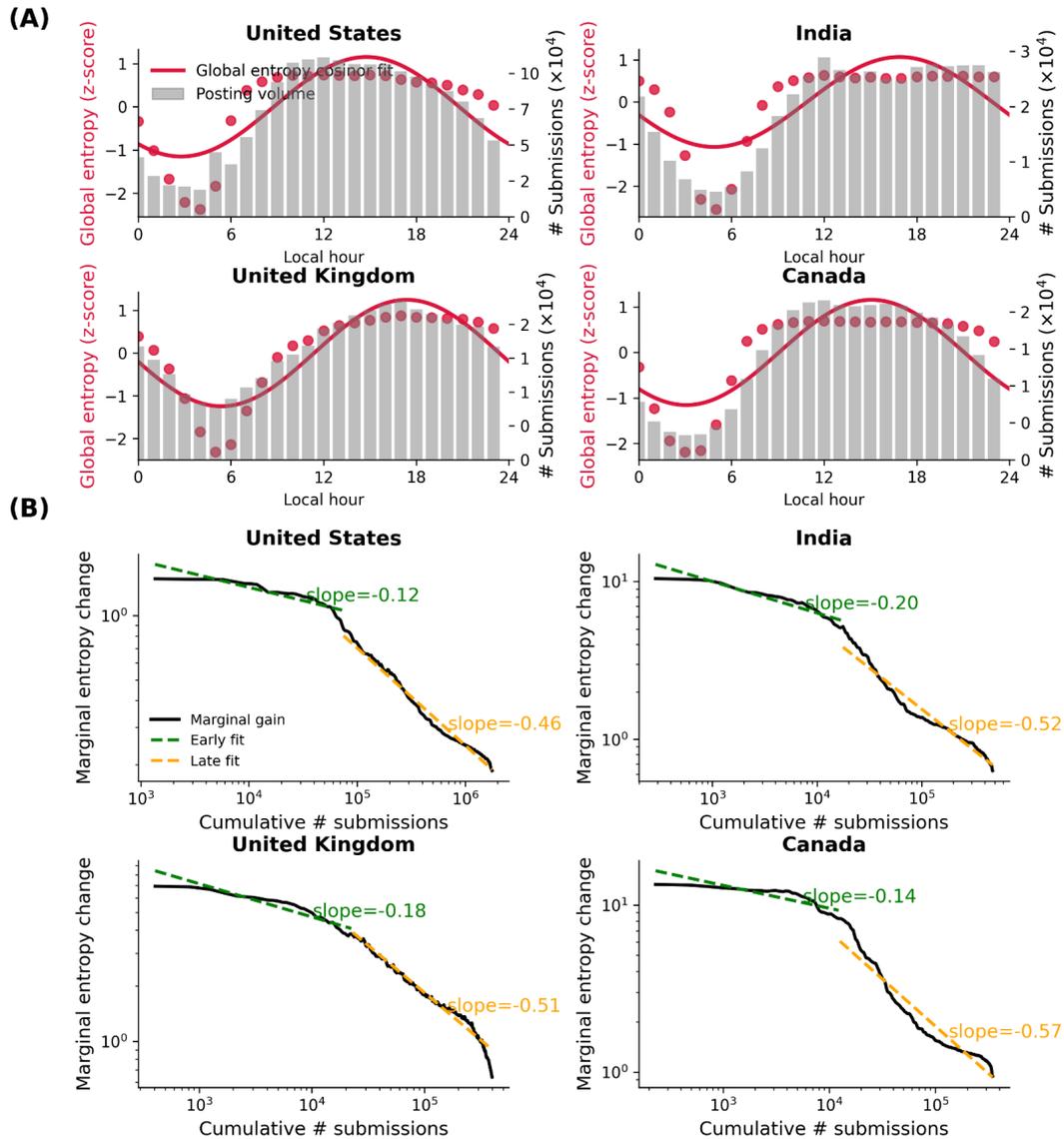

**Figure 5. Circadian global semantic entropy and diminishing returns of submission volume. (A)** Circadian profile of global semantic entropy (red circles) for four countries (United States, India, United Kingdom, Canada), z-scored within each country, with cosinor fits (red line) showing the 24-h rhythm. Gray bars indicate the total number of submissions per hour (scaled in ten thousands). **(B)** Cumulative contribution of individual posts to global semantic entropy for the same countries, calculated from the heatmaps in **Figure S3**. The marginal change in entropy per additional post (black line) exhibits strong diminishing returns: the latter segment (orange dashed line) shows a much steeper slope compared to the early segment (green dashed line). This pattern indicates that as posting volume increases, additional submissions contribute progressively less to overall semantic diversity, consistent with Zipf-Heaps-type constraints. Of note, with the x-axis and y-axis shown in log scale, the latter segment follows an approximate power-law scaling.

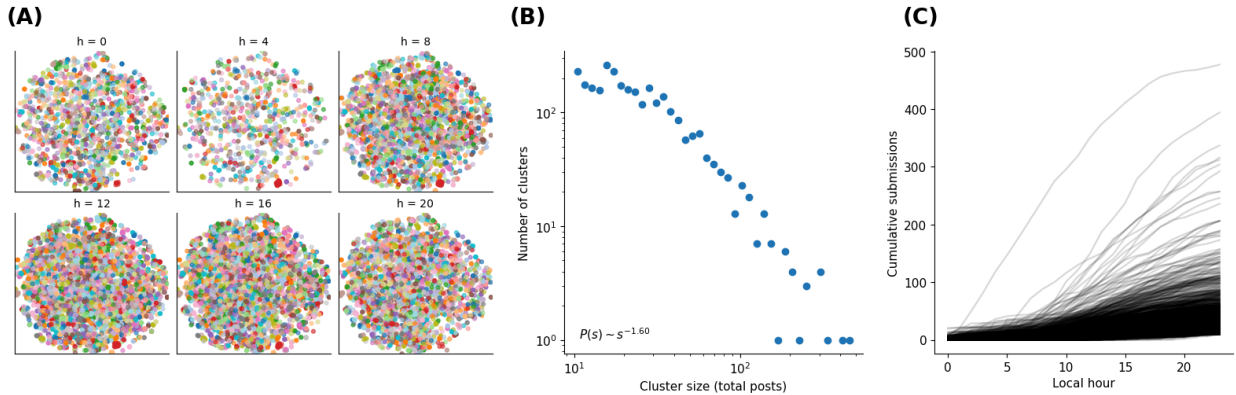

**Figure 6. Temporal organization and scaling of semantic clusters in the United States (January). (A)** Vectors of text embeddings were first reduced in dimensionality using PCA, then projected into two dimensions with t-SNE with PCA initialization (`scikit-learn`) for a stable starting configuration. Points are colored by `HDBSCAN` cluster assignment. Each subplot shows submissions from a specific local hour (sampled every 4 h). Early in the day, the semantic space exhibits widespread exploration, with posts distributed across multiple regions. Later in the day, new submissions predominantly accumulate within already established clusters, indicating continued activity within existing semantic topics rather than expansion into novel ones. Axes are fixed across subplots to enable direct comparison of spatial coverage across hours. **(B)** Log-log distribution of final cluster sizes (total number of submissions per cluster across the full day), revealing scale-free organization of semantic clusters. The fitted slope indicates an approximate power-law relationship between cluster size and frequency. **(C)** Cumulative growth of semantic clusters across the full 24-h cycle. Each curve represents the cumulative number of submissions within a single cluster over local time. Consistent with the power-law distribution in (B), the majority of submissions accumulate within a small number of large clusters.

## 4. Discussion

### 4.1. Cross-national consistency and seasonal entrainment of semantic exploration

Across four countries spanning different continents, cultures, and time zones, we observed a robust and highly consistent circadian rhythm in local semantic entropy. This cross-national convergence suggests that circadian modulation of semantic exploration reflects a general property of human cognitive dynamics rather than country-specific social schedules. Indeed, studies of mood in Twitter data similarly report a morning rise in positive affect across the globe. For example, Golder & Macy (2011) showed that positive mood has a morning rise and nighttime peak across different countries, and that this pattern persists even during weekends.

Importantly, human circadian rhythms are not only synchronized to the 24-h light-dark cycle but are also seasonally entrained by photoperiod. Seasonal variation in day length modulates circadian period and phase, producing systematic shifts in both molecular clock dynamics and overt behavioral rhythms (Myung et al., 2012; 2015). In line with this framework, we observed a positive correlation between sunset timing and the trough of local semantic entropy in the US, and complementary seasonal structure in the circadian heatmaps. Together, these findings suggest that circadian semantic exploration is sensitive to environmental light cues and varies across seasons, rather than being driven solely by local social timing.

### 4.2. Dissociating semantic exploration from sentiment and affect

A common hypothesis is that linguistic exploration may be secondary to mood, such that people express themselves more freely when feeling emotionally positive. We directly tested this account by examining the relationship between semantic entropy and sentiment measures. Contrary to a simple affect-driven explanation, correlations between semantic entropy and compound sentiment scores were weak, inconsistent in sign across countries, and insufficient to explain the observed circadian structure. In other words, periods of greater semantic diversity did not reliably coincide with more positive or negative affect.

This dissociation was further supported by comparisons with previously reported word-usage circadian rhythms. Local semantic entropy covaried with the temporal variability of sadness-related word usage extracted from Dzogang et al. (2017), but not with the variability of positive emotion (posemo) words. This selective alignment suggests that semantic exploration captures a dimension of cognitive behavior that is not reducible to affective valence or mood intensity.

From a neurobiological perspective, this pattern is consistent with evidence that dopamine primarily regulates motivation, incentive salience, and behavioral flexibility rather than hedonic experience per se (Berridge, 2018). Elevated dopaminergic tone promotes exploration of alternatives and increased variability in action selection without necessarily inducing positive affect. Our findings, therefore, suggest that semantic diversity reflects a form of cognitive exploration or flexibility that is partially independent of emotional state.

Interestingly, the sustained reduction in sentiment observed in the UK during early 2024 (January to May) likely reflects a convergence of seasonal affective factors (Bodden et al., 2024), economic fatigue, and political uncertainty (Harrois et al., 2025) rather than a single exogenous shock or major event. On the other hand, Canada exhibited consistently lower sentiment across the year, suggesting a structural baseline shift potentially driven by high-latitude seasonality, harsh weather (Mekis et al., 2015), and a linguistic tendency toward neutral or restrained expression (Machino et al., 2025).

### 4.3. Local and global semantic entropy reveals a two-stage daily process

A key contribution of this work is the distinction between local semantic entropy (diversity within a cluster) and global semantic entropy (diversity across the whole embedding space). Local semantic entropy showed that language use was most semantically diverse during the morning. In contrast, global semantic entropy peaked later in the day, closely tracking overall posting volume. This temporal dissociation reflects semantic exploration operating across scales. Early contributions introduce a wide range of semantic content, while later activity increasingly concentrates around these existing regions as submissions grow.

In fact, the non-linear relationship between global semantic entropy and submission volume can be understood in the context of well-established statistical patterns of language, notably classical Zipf's and Heaps' laws. Zipf's law predicts that word frequencies are highly skewed, with a small number of tokens dominating usage as corpus size increases (Ferrer i Cancho & Solé, 2003; Zipf, 1949), while Heaps' law predicts sublinear growth of vocabulary size with corpus size (Herdan, 1960). Consistent with these constraints, we find that increases in submission volume primarily amplify redundancy rather than novelty, as evidenced by a more than half reduction in the marginal contribution of additional posts to global semantic entropy from early to late activity regimes, indicative of Pareto-type diminishing returns, with the later regime exhibiting power-law scaling. Importantly, this pattern is not well described as a purely lexical scaling law, but is more naturally interpreted through a network-science perspective, following "rich-get-richer" dynamics akin to preferential attachment in growing networks (Barabási & Albert, 1999). Popular ideas and topics that already attract high visibility are more likely to receive further engagement and imitation, leading new contributions to cluster around existing semantic hubs rather than explore novel regions of semantic space.

### 4.4. Neurobiological rhythms underlying circadian semantic exploration

What biological mechanisms could drive this pattern? Neuromodulatory systems, and dopamine (DA) in particular, provide a plausible substrate. Although direct measurements of circadian dopamine rhythms in the human brain are limited, converging indirect evidence suggests higher dopaminergic activity during the daytime active phase and reduced activity during sleep. Retinal dopamine concentrations are higher in individuals who died during the daytime compared to nighttime (Korshunov et al., 2017), and a PET study measuring dopamine D2-receptor binding suggests reduced D2-receptor binding at night (Cervenka et al., 2008). A classic study found that plasma dopamine and norepinephrine peaks occur in the awake state, with markedly fewer DA peaks and lower mean levels during sleep (Sowers & Vlachakis, 1984). Together, these findings imply that dopaminergic tone in humans is elevated during the daytime and reduced during the nighttime. This pattern is consistent with our observation that local semantic entropy, reflecting individual-level semantic exploration, peaks in the morning.

In contrast, extensive work in rodents, which are nocturnal animals, consistently reports dopamine rhythms that peak during the dark, active phase. In freely moving rats and mice, extracellular dopamine in the striatum is higher at night than during the light phase (Smith et al., 1992). Optical dopamine sensors further demonstrate that striatal dopamine is highest during wakefulness and lowest during REM sleep in mice (Dong et al., 2019), directly linking DA levels to the sleep-wake cycle. The nucleus accumbens in rats also exhibits a clear circadian DA rhythm that persists under constant light or dark, indicating an intrinsic clock mechanism (Castañeda et al., 2004). Thus, dopaminergic rhythms in rodents are robust but antiphasic to those inferred in humans.

This species difference highlights the importance of considering diurnality versus nocturnality when linking neuromodulatory mechanisms to behavior. Across species, dopaminergic signaling has been tightly linked to exploratory behavior, particularly through its role in regulating variability in action selection and the exploration-exploitation trade-off (Humphries et al., 2012; Markowitz et al., 2023). Mechanistically, the neuromodulatory system has been linked to fluctuations in neural gain, arousal, and population-level neural variability, which are factors that influence cognitive flexibility (McGinley et al., 2015; Shine et al., 2018). If semantic exploration in language reflects domain-general mechanisms of cognitive flexibility, then daily modulation of dopaminergic tone provides a biologically plausible substrate for the circadian structure we observe in semantic entropy.

Taken together, these findings extend circadian analyses of language beyond word frequency and affect to the structure of semantic space itself. The consistent early-day peak in local semantic entropy and the delayed peak in global semantic entropy across countries suggest that circadian modulation influences not only what people say, but how diversely they express ideas. This pattern aligns with broader evidence that cognitive flexibility, variability, and exploratory behavior fluctuate across the day, supporting the view that circadian biology shapes fundamental properties of human cognition in naturalistic settings.

**Acknowledgements**

This work was partially supported by Taipei Medical University, Office of Global Engagement, through the International Collaborative Research Projects Grant (DP3-112-52322); the Higher Education Sprout Project of the Ministry of Education (MOE) in Taiwan; and the National Science and Technology Council (NSTC), Taiwan (113-2314-B-038-121, 114-2320-B-038-052-MY3).

**Data availability**

The raw Reddit data analyzed in this study are available via the Pushshift archive. Due to copyright and privacy considerations, we do not redistribute raw text. Preprocessed data and analysis scripts necessary to reproduce the reported results can be accessed at our GitHub repository: `https://github.com/vuonghtruong/reddit-semantic-entropy`. Additional materials or data are available from the authors upon reasonable request.

# Supplementary Figures

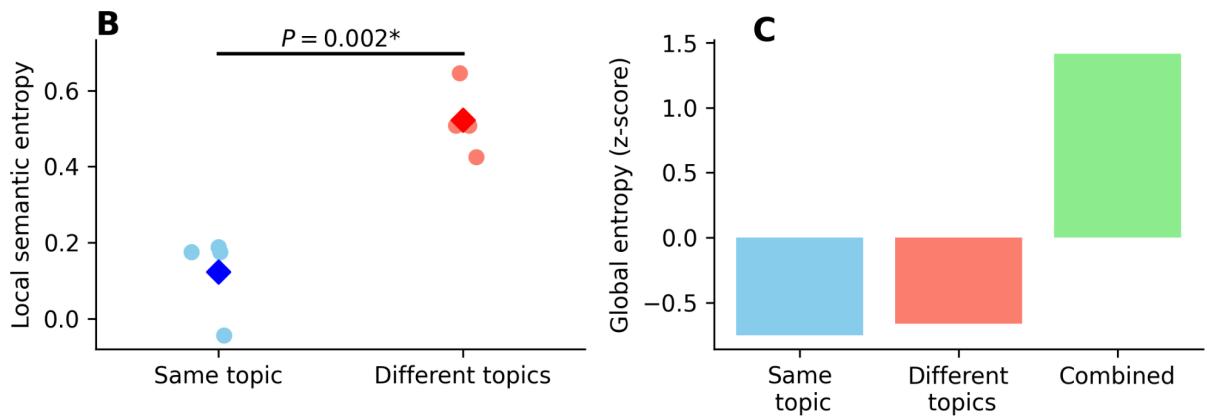

**Figure S1. Conceptual illustration and validation of local and global semantic entropy**. (**A**) Example text sets illustrating low versus high semantic diversity. Left: sentences drawn from the same topic (sports), yielding low semantic dispersion. Right: sentences drawn from different topics, yielding higher semantic dispersion in embedding space. (**B**) Local semantic entropy computed for the same-topic and different-topic text sets. Points represent individual samples; diamonds indicate group means. Different-topic sets exhibit significantly higher local semantic entropy than same-topic sets (two-sample $t$ test, $P = 0.002$). (**C**) Global semantic entropy (z-scored) for same-topic, different-topic, and combined (same + different) text sets. While same-topic and different-topic sets alone show relatively low global entropy, combining them markedly increases global entropy, indicating that global entropy reflects population-level dispersion across the full semantic space rather than within-set variability.

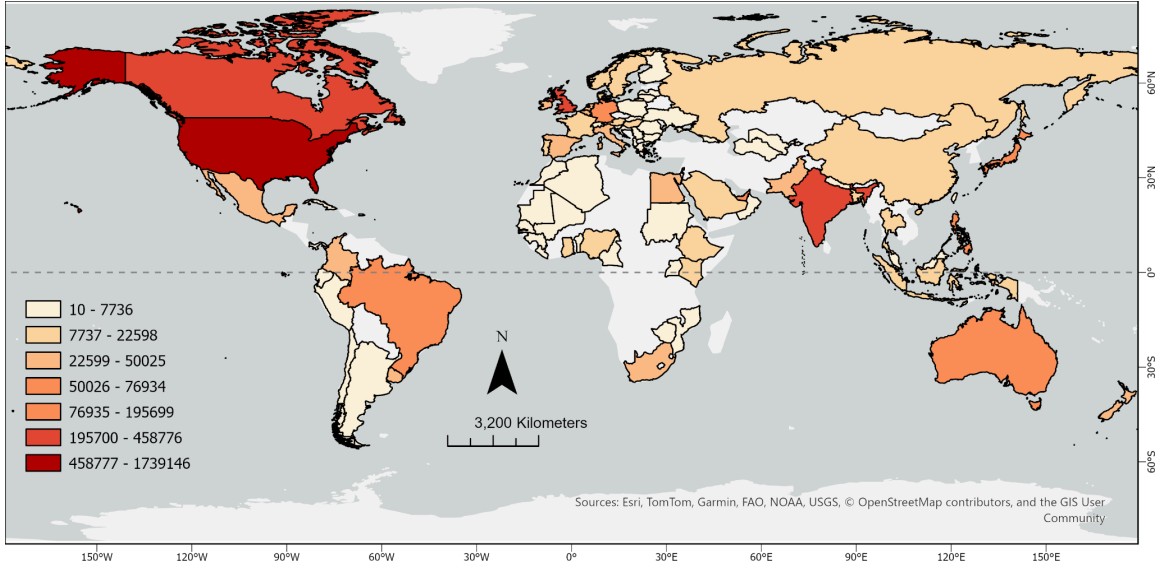

**Figure S2. Global distribution of Reddit submissions.** World map showing the total number of English-language Reddit submissions aggregated by country from January 2024 to December 2024. Color intensity indicates submission counts. The gray dashed line denotes the equator, shown for geographic reference.

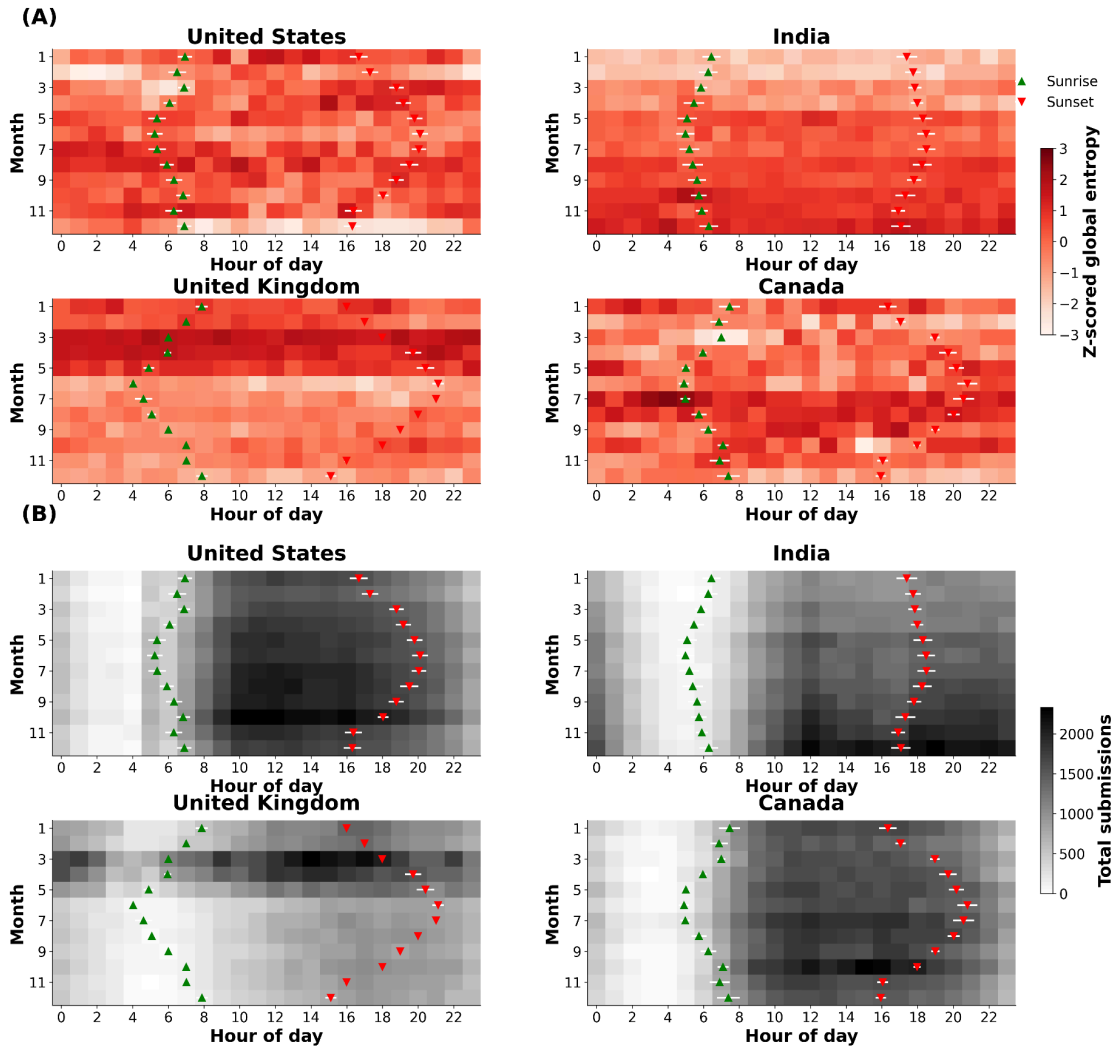

**Figure S3.** Circadian-seasonal heatmaps of global semantic entropy and total submission activity across countries. (**A**) Heatmaps of z-scored global semantic entropy by month (rows) and local hour (columns) for the US, India, UK, and Canada. Warmer colors indicate higher semantic entropy. Green triangles denote mean sunrise times ± SD, and red inverted triangles denote mean sunset times ± SD. (**B**) Heatmaps of total number of submissions per month and hour for the same countries, showing periods of higher posting activity. Sunrise and sunset times are overlaid similarly as in panel (A).